# SPOOK: A system for probabilistic object-oriented knowledge representation

Avi Pfeffer     Daphne Koller     Brian Milch     Ken T. Takusagawa
Computer Science Department
Stanford University
{avi,koller,milch,kenta}@cs.stanford.edu

## Abstract

In previous work, we pointed out the limitations of standard Bayesian networks as a modeling framework for large, complex domains. We proposed a new, richly structured modeling language, *Object-oriented Bayesian Networks*, that we argued would be able to deal with such domains. However, it turns out that OOBNs are not expressive enough to model many interesting aspects of complex domains: the existence of specific named objects, arbitrary relations between objects, and uncertainty over domain structure. These aspects are crucial in real-world domains such as battlefield awareness. In this paper, we present SPOOK, an implemented system that addresses these limitations. SPOOK implements a more expressive language that allows it to represent the battlespace domain naturally and compactly. We present a new inference algorithm that utilizes the model structure in a fundamental way, and show empirically that it achieves orders of magnitude speedup over existing approaches.

## 1 Introduction

Bayesian networks are a graphical representation language in which complex probabilistic models can be represented compactly and naturally. The power of the representation comes from its ability to capture certain structure in the domain — the locality of influence among different attributes. This structure, which is formalized as probabilistic conditional independence, is the key to the compact representation. It also supports effective inference algorithms.

In previous work [KP97], we argued that, despite their power, Bayesian networks (BNs) are not adequate for dealing with large complex domains. Such domains require an explicit representation of additional types of structure: the notion of an *object*, a complex structured domain entity with its own properties; and the notion of a *class* of objects, that captures properties common to an entire set of similar objects. Our *Object-Oriented Bayesian Networks (OOBNs)* extended the language of BNs with these additional concepts.

By introducing objects and classes, OOBNs provide us with a representation language that makes it much easier to specify large models in a compact and modular way. However, these new concepts also reveal the shortcomings of the OOBN framework. As soon as we have objects, we want to encode various relationships between them that go beyond part-whole. For example, we may have an object representing some physical location (with its own properties). We may well wish to assert that another object, such as a military unit, is *at* the location. This relation is not a part-whole relation, and thus does not fit naturally into the OOBN framework.

In [KP98], we described a language that allows a much richer web of relations between objects. It also extends the expressive power of the language in several significant ways. For example, by making relations first-class citizens in our ontology, we can express knowledge we might have about them; just as importantly, we can express *lack of knowledge* about relations. For example, we can express the fact that we do not know which of several locations a unit is at; we can even quantify this uncertainty using probabilities. We can also express uncertainty about the number of subunits that a unit has.

Although the additional expressive power provided by OOBNs and its extensions is natural and even desirable, one still needs to make the case that it actually helps model real-life domains. We need also to show that we have the capability to answer interesting queries using a reasonable amount of computation. In this paper, we address both of these points. We present an implemented system called SPOOK — System for Probabilistic Object-Oriented Knowledge. We show that it can be used to represent and reason about a real-world complex domain.

The domain we have chosen for this test is military situation assessment [ML96]. This domain is notoriously challenging for traditional Bayesian networks. It involves a large number of objects, related to each other in a variety of ways. There is also a lot of variability in the models appropriate to different situations. We started with a set of Bayesian networks constructed for this domain by IET, Inc.

We then used our SPOOK language to construct a single unified model for this domain, one with a rich class hierarchy. The resulting model was compact, modular, natural, and easy to build.

We also investigate our ability to answer queries effectively using such a complex model. One approach (the one we proposed in [KP98]) is based on *knowledge based model construction (KBMC)* [WBG92] — converting the complex model into a traditional BN, and using standard BN inference. The BNs constructed from a complex SPOOK model are large enough to stretch the limitations of existing inference algorithms. Even the network for a single SCUD battalion involves over 1000 nodes and requires 20 minutes to answer a query. A network for many interacting units in a battlespace would be orders of magnitude larger.

The challenges posed by real-life complex models require a more sophisticated approach to inference. In our original OOBN paper [KP97], we described an inference algorithm that makes use of the structure made explicit in the more expressive language: the encapsulation of one object within another, and the model reuse implied by the class hierarchy. The OOBN algorithm is too simple to apply to our much richer SPOOK language. However, it turns out that many of the same ideas can be adapted to this task. We present a new inference algorithm, implemented in the system, that utilizes encapsulation and reuse in a fundamental way. We present experimental results for our algorithm on realistic queries over the battlespace model, and show that by utilizing encapsulation and model reuse, we obtain orders of magnitude speedup over the KBMC approach.

## 2 The SPOOK Language

In this section we review the SPOOK representation language. The language presented here is based on the probabilistic frame systems of [KP98]; it extends the language of object-oriented Bayesian networks (OOBNs) [KP97] in several important ways.

The basic unit in the SPOOK language is an *object*. An object has *attributes*, which may be either *simple* or *complex*. A simple attribute is a function from objects to values in some specified domain; it is similar to a variable in a standard BN. A complex attribute represents a relationship between objects. If the value of complex attribute $A$ of object $X$ is $Y$ (notated $X.A = Y$), the relation $A(X, Y)$ holds. Complex attributes may be *single-valued*, corresponding to functional relationships, or *multi-valued*, corresponding to general binary relations. A complex attribute may have an *inverse*: if the inverse of attribute $A$ is $B$, and $Y$ is a value of $X.A$, then $X$ must be a value of $Y.B$.

For example, a scud-battalion object has a simple attribute under-fire, whose value ranges over {none, light, heavy}. It has a single-valued complex attribute at-location, whose value is an object corresponding to the location of the battalion. It has a multi-valued complex attribute has-battery, each of whose values is a battery in the battalion. The has-battery attribute has an inverse in-battalion, which is a single-valued complex attribute of a battery object. If battery-1 is a value of scud-battalion-charlie.has-battery, then battery-1.in-battalion = scud-battalion-charlie. The dot notation can be extended to *attribute chains* $A_1.A_2.\cdots.A_k$, denoting the composition of the relations $A_1, \ldots, A_k$. If $A_1, \ldots, A_{k-1}$ are single-valued complex attributes, and $A_k$ is a simple attribute, we call the attribute chain *simple*.

The probability model for an object is specified by defining a local probability model for each of its simple attributes. As in BNs, The local probability model consists of a set of parents, and a conditional probability distribution (CPD). A parent can be either another simple attribute of the same object, or a simple attribute chain. Allowing attribute chains as parents provides a way for the attributes of an object to be influenced probabilistically by attributes of related objects. If two objects are inverses of each other, each can be influenced by the other.

Continuing our example, the under-fire attribute of scud-battalion has a parent at-location.defense-support, and the CPD for under-fire indicates that the battalion is more likely to be under heavy fire if it is in a location with poor defense support. The battery object has a hit attribute whose parent is in-battalion.under-fire, thus creating an indirect chain of influence from the location, through the battalion at the location, to the battery in the battalion. Since in-battalion is an inverse of has-battery, the battalion can in turn be influenced by the battery it contains. For example the attribute scud-battalion.next-activity depends on has-battery.launch-capability. (Section 2.2 explains how to specify dependence on multi-valued attributes.)

### 2.1 Classes and instances

In SPOOK, a probability model is associated with a *class*, which corresponds to a type of entity in the domain. An *instance* of a class corresponds to a domain entity of the appropriate type, and derives its probability model from its class. We use *object* to denote either a class or an instance. For example, scud-battalion is a class and scud-battalion-charlie is an instance of scud-battalion.

Classes provide reusable probability models, that can be applied to many different objects. Classes are organized in a *class hierarchy*. A *subclass* inherits the probability model of its *superclass*, and it can also override or extend it. The inheritance mechanism facilitates model reuse by allowing the commonalities between different classes to be captured in a common superclass. For example, the battalion superclass captures those features common to all battalions.

Classes also provide a type system for the SPOOK language.

Every complex attribute $A$ has a type $T(A)$, and for any object $X$, the value of $X.A$ must be an instance of $T(A)$. If no particular value is specified for $X.A$, we use the *unique names assumption*, which states that the values of $X.A$ are generic, unnamed instances of $T(A)$, that are not related in any other way to the instances in the model.

The unique names assumption implies that in the class models, no two battalions can be at the same location. Instances provide a way to specify such webs of inter-related objects. In this example, there are two battalion instances, battalion-1 and battalion-2, and a location instance location-a. By stating that battalion-1.at-location = location-a and that battalion-2.at-location = location-a, the objects are hooked together appropriately.

### 2.2 Multi-valued attributes & structural uncertainty

As discussed above, a complex attribute can be multi-valued, but a parent of a simple attribute must be a simple attribute chain, in which the attributes are single-valued. In order to allow the attributes of an object to be influenced by attributes of related objects when the relationship is multi-valued, we introduce a *quantifier attribute*. A quantifier attribute has the form $\#(A.\rho = v)$, where $A$ is a multi-valued complex attribute, $\rho$ is a simple attribute chain, and $v$ is a possible value of $\rho$. If $X$ is an object with attribute $A$, $X.\#(A.\rho = v)$ denotes the number of objects $Y$ such that $A(X, Y) \wedge Y.\rho = v$.

Quantifier attributes allow attributes of an object to depend on aggregate properties of a set of related objects. Continuing our running example, we may specify that a parent of scud-battalion.next-mission is the quantifier attribute #(has-battery.launch-capability=*high*). The value of the quantifier is determined by the value of launch-capability for each of the batteries in the battalion. If the set of batteries in the battalion is fixed, the quantifier simply expresses an aggregate property of the set. However, we may also have uncertainty over the number of batteries in the battalion. This is an example of *structural uncertainty*, which is uncertainty not only over the properties of objects in the model but over the relational structure of the model itself.

The type of structural uncertainty encountered in this example is *number uncertainty*: uncertainty over the number of values of a multi-valued complex attribute. Number uncertainty is integrated directly into the probability model of an object using a *number attribute*. If $A$ is a multi-valued complex slot, the number attribute $\#A$ denotes the number of values of $A$. A number attribute is a standard random variable whose range is the set of integers from 0 to some upper bound $n$. It can participate directly in the probability model like any other variable. In our example, scud-battalion.#has-battery depends probabilistically on scud-battalion.country. Under number uncertainty, the value of a quantifier depends on the value of the number attribute, as well as on the values of the related objects.

Another kind of structural uncertainty is *reference uncertainty*, which is uncertainty over the value of a single-valued complex attribute. For example, we may have uncertainty over whether a battalion is located in a mountain or a desert location. As with number uncertainty, reference uncertainty can be introduced directly into the probability model of an object using a *reference attribute*. If $A$ is a single-valued complex attribute whose value is uncertain, $\mathcal{R}(A)$ is a reference attribute whose range determines the possible values of $A$. An element of the range of $\mathcal{R}(A)$ may either be a subclass $C$ of $T(A)$, or an instance $I$ of $T(A)$. If the value of $X.\mathcal{R}(A)$ is the type $C$, then the value of $X.A$ is a generic instance of $C$; if the value of $X.\mathcal{R}(A)$ is the instance $I$, then the value of $X.A$ is $I$. As with number attributes, reference attributes participate in the probability model, and can depend on and be influenced by other attributes. We call this type of uncertainty "reference uncertainty" because we do not know which object is being referred to when we refer to the value of $A$.

A SPOOK knowledge base consists of a set of classes and instance models. In [KP98], we defined a data structure called a *dependency graph* that can be used to make sure that all the probabilistic influences, including the influences between different objects, are acyclic. We defined a semantics for SPOOK models, based on a generative process that randomly generates values for attributes of instances in the domain, including number and reference attributes. We showed that if the dependency graph is acyclic, then the knowledge base defines a unique probability distribution over the values of all simple attributes of all named instances in the KB.

## 3 Modeling the Battlespace Domain

To demonstrate the representational power of the SPOOK language, we implemented a model for reasoning about military units in a battlespace. In [LM97], Mahoney and Laskey describe how they model this domain using *network fragments*. In this section, we introduce the domain, discuss why it is difficult to model using BNs, and describe how we modeled it using SPOOK.

The purpose of the battlespace model is to reason about the locations and status of military units based on intelligence reports. Our model deals specifically with missile battalions, the batteries within those battalions, and the individual units — vehicles, radar emplacements, missile launchers, etc. — within the batteries. A scenario consists of multiple battalions, some of which may be at the same location. A battalion typically has four batteries, each with about 50 units. Thus, the model for a battalion includes about 200 units, and a scenario may include 1000 units.

Let us consider trying to model our domain directly with a BN. With four or five variables for each unit, a flat BN for a battalion model will typically contain over a thousand nodes. The sheer size of this network is a major obstacle to its construction. In addition, the resulting BN will be too rigid for practical purposes. The configuration of a battalion is highly flexible, with the exact number of units of each type varying considerably between different battalions.

These difficulties have led to an alternative approach, in which several different BNs are used, one for each aspect of the model. Figure 1(a) shows a Bayesian network for an SA3 battalion. There are similar networks for other types of units, such as Scud battalions and batteries. Although a Scud battalion contains Scud batteries, the battalion model does not replicate all the details of the battery model; rather, it summarizes the status of all the batteries with nodes, indicating the initial number of batteries, the number of damaged batteries, and the current number. These summaries serve two purposes: to keep the network reasonably simple; and to account for changing model configuration by making the initial number of subunits a variable.

A major disadvantage of this approach is that it is very difficult to reason between the different networks. The only way to reason from one network to another is to reach conclusions about the state of variables in one network and assert them as evidence in the other network. For example, the only way to transfer conclusions from a battery to a battalion is to condition one of the summary nodes in the battalion model; going from one battery to another requires conditioning the battalion model, reasoning about the battalion, and then conditioning the other battery model. This type of reasoning has no sound probabilistic semantics. There is no way to combine evidence about multiple different units in a probabilistically coherent manner. Furthermore, this type of reasoning between fragments must be performed by a human or a program. It requires some model of the relationship between the fragments, e.g., that the status node of the battery model is related to the number-damaged-batteries node of the battalion model. Nowhere is this relationship made explicit.

Another disadvantage is that multiple BNs do not allow us to take advantage of redundancy within a model and similarities between models. For example, the battalion model in Figure 1(a) contains many similar substructures, summarizing groups of units of different kinds. In addition, different battalions may all have substructures describing their locations, as shown in the bottom right corner of the figure. In the multiple BNs approach, the only mechanism for exploiting these redundancies is cut-and-paste. This makes it very hard to maintain these models, because each time one of the reused components is changed, it must be updated in all the different networks that use it.

OOBNs solve the problems inherent in the multiple BN approach. By allowing a battalion to contain a battery as a sub-object, we can easily have the battalion model encompass the complete models of the different batteries in it, which in turn contain complete models of their subunits, without making the battalion model impossibly complex. We can then reason between different objects in the part-of hierarchy in a probabilistically coherent manner. In addition, by allowing us to define a class hierarchy, OOBNs allow us to exploit the redundancy in the model.

However, the language of OOBNs is quite restricted, in a way that is problematic in our domain. If we want to model the effect of a unit's location on the unit, we need to represent the relationship between the unit and its location. In our model, this was the only relationship that did not fall into the part-of hierarchy, but richer models of the battlespace domain require more sophisticated relationships, such as that between a unit supporting another unit. In addition, our domain requires multi-valued attributes and quantifiers. A battalion contains several batteries, and each battery contains several units of different types. The higher level objects do not depend directly on the individual lower level objects, but only on aggregate properties of the set of objects, expressed through quantifier attributes. The ability to create named instances and hook them together via relations is also important in our domain, as illustrated by the example from the previous section of two battalions in the same location. Finally, the battlespace domain contains a great deal of structural uncertainty, in particular number uncertainty over the number of subunits. One may also have reference uncertainty as to the location of a unit.

SPOOK includes all the capabilities of OOBNs to represent part-of and class hierarchies, and also handles relations between objects, multi-valued attributes, named instances, and structural uncertainty, all of which cannot be expressed in OOBNs. Our SPOOK model of the battlespace domain includes a natural class hierarchy, with Military-Unit, Environment, Location and Weather as root classes. The Battalion, Battery, Group, and Unit families are all part of the Military-Unit hierarchy. Similarly, part-of relationships are easy to model in SPOOK using inverse relations. The has-battery attribute of a battalion, and the in-battalion attribute of a battery, are inverses, allowing the battalion and its contained battery to influence each other. Batteries do not contain individual units directly, but instead contain a Group object for each type of unit. For instance, a battery has (among others) groups of missile launchers, command vehicles, and anti-aircraft artillery units. Each Group has a multi-valued attribute relating it to the individual units, as well as a number attribute and a set of quantifier attributes that summarize the status of the units. Using Group objects is convenient because we summarize the same attributes for all types of units.

An object of class Unit has simple attributes reported, operational, damaged and reported-damaged. These attributes are influenced by the location of the battalion — specifi-cally, the location's support for concealment and defense — and by the battalion being under fire. We represent these influences in SPOOK by specifying, for example, at-location.defense-support as a parent of damaged. The number of damaged units in turn influences the battery's operational attribute, and a quantifier slot that counts the number of operational batteries in a battalion influences the battalion's current-activity. Subclassing gives us the ability to provide models for certain types of units that are similar to the general unit model but not exactly the same. For instance, Missile-Launcher has an additional activity attribute that indicates whether it is launching, reloading, or idle. While we only modeled the domain up to the battalion level, we could easily extend our model to higher-level groups in the military hierarchy.

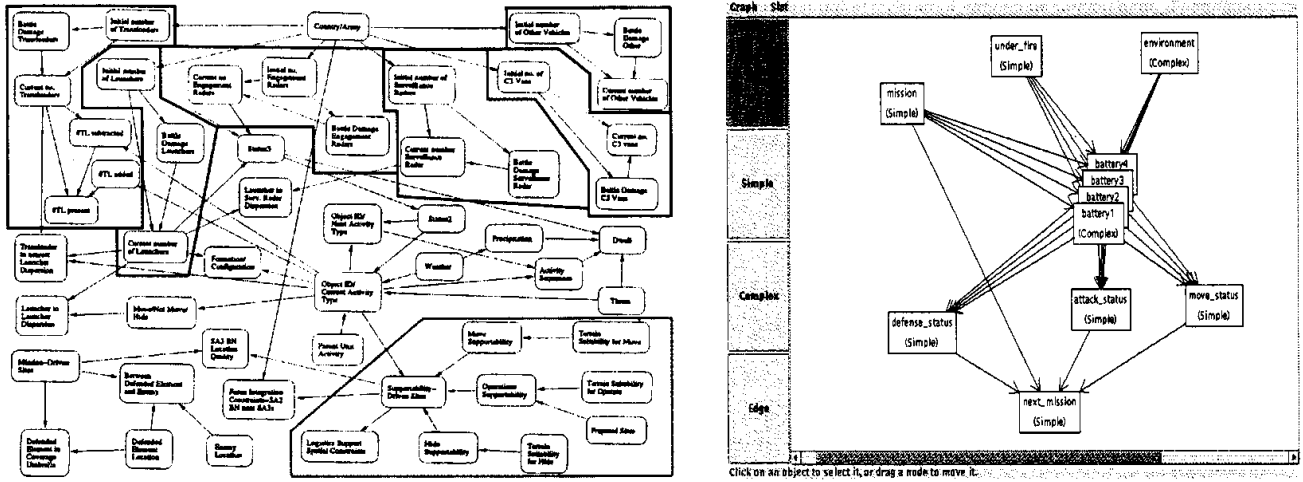

Figure 1: (a) SA3 Battalion Bayesian network, (b) SPOOK model of Scud Battalion

In our current model, all units in a battalion share a common environment, which is referred to by the in-environment attribute of the battalion. The environment is composed of Location and Weather objects, which between them determine the current support of the environment for various activities such as moving, hiding and launching missiles. We could have associated a different environment with each battery or unit, making locations of lower-level objects related probabilistically to higher level objects.

To give an example of the power of reasoning at multiple levels of the hierarchy and between different objects, we present a series of queries we asked the model. First we queried the prior probability that a particular Scud battery was hit, and found it to be 0.06. We then observed that the containing battalion was under heavy fire, and the probability that the battery was hit went up to 0.44. We then observed, however, that none of the launchers in the battery had been reported to be damaged, and the probability that the battery was hit went down to 0.28. We then explained away this last observation, by observing that the environment has good support for hiding; the probability that the battery was hit went back up to 0.33. This example combines causal, evidential and intercausal reasoning, and involves battery and battalion objects, individual launcher objects, the launcher group, and the environment object.

## 4 Inference

In the previous sections we described the SPOOK language, and how we used it to model the battlespace awareness domain. Of course, in order for the language to be useful, we need an efficient algorithm for performing inference in it. Ideally, we would like the language features to lend themselves to efficient inference. Indeed, as we argued in [KP97], one of the measures of a successful representation language is that it makes the structure of the domain explicit, so that it can be exploited by an appropriately designed inference algorithm.

One way to perform inference in SPOOK is to use the technique of *knowledge-based model construction (KBMC)* [WBG92]. In this approach, we construct a BN that represents the same probability distribution as that defined by our model, and answer queries in this BN using standard BN inference algorithms. We described the KBMC process for our language in detail in [KP98], and showed that if the dependency graph is acyclic, it always terminates.

While the KBMC approach provides a sound and complete method for answering queries in SPOOK it is somewhat unsatisfactory. It fails to exploit the language's ability to make explicit the structure of the domain. All notions of objects and relationships are lost when the model is turned into a flat BN. In [KP97], we argued that the object structure of a model can and should be exploited for efficient inference. We argued that two aspects of the structure in particular should be exploited: the fact that the internal state of each object is encapsulated from the remainder of the model via its interface, and that locality of interaction typically induces small interfaces; and the fact that the same type of object may appear many times in a model. Since the flat BN produced by KBMC has no notion of an object, the KBMC algorithm cannot exploit these structural features.

We now present an object-based inference algorithm that does exploit the structural features of a model. The algorithm is based on the ideas presented in [KP97], but it is significantly more complex due to the increased expressivity of our language. The added complexity arises principally from four new language features.

First, the "multi-centeredness" of the language implies that each object can be accessed by a multitude of other objects, in a variety of different ways. In OOBNs, we assumed that each type of object had a unique set of inputs and outputs, and that we could precompute a conditional distribution over the outputs given the inputs. This is no longer the case. Because an object can be accessed in many different ways, its outputs can be arbitrarily complex. In addition, its inputs are not fixed, but are determined by the way the object is accessed, and the particular set of outputs required, as will be explained below. Thus, for each object referred to by another object, our algorithm must determine its inputs and outputs on the fly, during the processing of a query.

The second relevant language feature is the ability to create instances and hook instances together via relations. As we shall explain later, this property implies that encapsulation, although still present, no longer holds in exactly the same way as in OOBNs. The third feature is multi-valued attributes and quantifier attributes that depend on them, which do not appear in OOBNs, and require a new treatment.

The final complicating feature is structural uncertainty. The naive approach to dealing with structural uncertainty, that could be applied to OOBN models, is as follows. To compute $P(Q)$, enumerate all possible structural hypotheses $h$, and compute $P(Q \mid h)$ for each such hypothesis. $P(Q)$ is then equal to $\sum_h P(h)P(Q \mid h)$. Unfortunately, the number of structural hypotheses is exponential in the number of structural variables, rendering this approach completely infeasible with more than a very small number of structural variables. In the battlespace awareness domain, the number of structural variables is large, since we have uncertainty over the number of units in many different groups. Therefore we need a much better way of doing inference with structural uncertainty.

Our inference algorithm is related to the KBMC algorithms, but its recursive nature makes it quite different, so we describe it in detail. It is fairly complex, so we present it in stages. We begin with the basic algorithm, for class objects without multi-valued complex attributes, we then extend it to deal with instances, and finally we show how we deal with multi-valued attributes, quantifier attributes, number uncertainty and reference uncertainty.

### 4.1 Basic algorithm

Our inference algorithm is recursive. The main function of the algorithm answers a query on an object. The key to understanding the algorithm is to understand how the function is recursively called, and the relationship between the object calling the function and the object on which it is called. Suppose that, during the process of solving a query on an object $X$, we encounter a complex attribute $D$ of $X$. For now, let us assume that $D$ is single-valued. There is some object $Y$ that is the value of $X.D$. Let us assume for now that no value is asserted in the knowledge base for $X.D$, so that $Y$ is a generic (unnamed) instance of $T(D)$.

Recall that other attributes of $X$ may depend on the value of $D$, i.e., on $Y$. Specifically, let $\sigma_1, \ldots, \sigma_n$ be the complete set of attribute chains, such that attributes of $X$ depend on each of the $Y.\sigma_i$ but on no other aspects of the value of $Y$. In order to solve a query on $X$, we will need to solve a subquery on $Y$, to obtain a distribution over $Y.\sigma_1, \ldots, Y.\sigma_n$. Recall, however, that $Y$ may itself depend on $X$. This will happen if $D$ has an inverse $E$, so that the value of $Y.E$ is $X$. Let $\tau_1, \ldots, \tau_m$ be the complete set of attribute chains through which $Y$ depends on $X$. The distribution over $Y.\sigma_1, \ldots, Y.\sigma_n$ will depend on the values of $X.\tau_1, \ldots, X.\tau_m$. The subquery on $Y$ needs to return a conditional distribution over $\sigma_1, \ldots, \sigma_n$ given $\tau_1, \ldots, \tau_m$. The issue is further complicated by the fact that, while solving a query for object $X$, we do not yet know the set $\tau_1, \ldots, \tau_m$, through which $Y$ depends on $X$. This information can only be computed within $Y$ itself. Therefore, when answering the subquery on $Y$, we return not only the conditional distribution over $\sigma_1, \ldots, \sigma_n$, but also the conditioning variables $\tau_1, \ldots, \tau_m$.

The main function of our algorithm, **SolveQuery**, takes three arguments, one of which is optional. The two required arguments are an object $Y$, called the *target* of the query, and a set of attribute chains $\sigma = \sigma_1, \ldots, \sigma_n$, called the *outputs* of the query. The optional argument is an attribute $E$, called the *entry point* of the query; $E$ is the entry point into $Y$ if $Y.E$ is $X$. The entry point is used for discovering the dependencies of $Y$ on $X$: $Y$ depends on $X.\tau$ only when some attribute in $Y$ depends on $B.\tau$. In this case, $\tau$ is said to be an *input* to the query. **SolveQuery** returns two values: the set of inputs $\tau = \tau_1, \ldots, \tau_m$ to the query, and a conditional probability distribution over $\sigma$ given $\tau$. A query may have no entry point if it is the top-level call to **SolveQuery** or if it was called for an attribute $D$ of $X$ that has no inverse. In that case, $Y$ cannot get inputs from $X$, so that $\tau$ will be empty, and the distribution returned over $\sigma$ will be unconditional.

The basic procedure of **SolveQuery** is as follows.

**SolveQuery** constructs a local BN, which it will eventually use to solve the query. The BN consists of nodes for each of the attributes of the query target $Y$, nodes for the inputs and outputs, and other nodes that help communicate between different attributes. In order to add a node to the network, we must complete four steps: create it, specify its range, specify its parents, and specify its conditional probability distribution (CPD). These steps are not always performed together; in some cases, we cannot specify the CPD of a node at the time that it is created.

**SolveQuery** begins with an initialization phase. First it creates a node $\nu(A)$ in the network for every attribute $A$ of $Y$. For each simple attribute $A$, we specify the range of $\nu(A)$ to be the range of $A$. If $A$ is complex, we want the range to be the product of all the attribute chains through which $Y$ depends on $A$, but we do not yet know this set. For this reason, we maintain a set $needed(A)$ for each complex attribute $A$.

Next, **SolveQuery** creates an *output node* $\nu(output)$, to represent the query output. The range of $\nu(output)$ is $\times_{i=1}^{n} \text{Dom}(\sigma_i)$, where $\text{Dom}(\sigma_i)$ is the range of the simple attribute at the end of the attribute chain $\sigma_i$. For each $\sigma_i$, we call the function **GetChainNode** (see below) to obtain a node $\nu(\sigma_i)$, whose range is $\text{Dom}(\sigma_i)$, and make it a parent of $\nu(output)$. The CPD for $\nu(output)$ simply implements the cross-product operation: if the values of $\nu(\sigma_1), \ldots, \nu(\sigma_n)$ are $v_1, \ldots, v_n$, the value of $\nu(output)$ is $\langle v_1, \ldots, v_n \rangle$ with probability 1.

**GetChainNode** is called whenever we need to produce a node to represent the value of an attribute chain $\sigma$. If $\nu(\sigma)$ is already in the BN, we simply return it. This will always be the case if $\sigma$ is just a simple attribute $A$. Otherwise, $\sigma$ must have the form $A.\rho$, where $A$ is a complex attribute. The algorithm thus needs to ensure that the processing of $A$ will give the required information about $A.\rho$. We therefore add $\rho$ to the set $needed(A)$. We can extract the value from the output of $A$ by creating a new *projection node* $\nu(\sigma)$, whose range is $\text{Dom}(\sigma)$, and set its lone parent to be $\nu(A)$. As we will see below, the projection node performs the inverse operation to that of the cross-product node.

The main phase of **SolveQuery** performs a backward-chaining process to determine the interfaces of complex attributes. First, we order the attributes of $Y$ in an order consistent with the dependency graph. Such an order must exist if the model is well-defined. We then process the attributes one by one, in a bottom-up order. Children must precede their parents, since processing a child tells us what "information" we need from its parents. Processing a simple attribute $A$ is easy. We simply obtain the set of attribute chain parents of $A$, as specified in the model of $Y$. For each such parent $\sigma$, we convert it into a BN node $\nu(\sigma)$ by calling **GetChainNode**, and add it as a parent of $\nu(A)$. We then set the CPD of $\nu(A)$ as specified in the model of $Y$.

Processing a complex attribute $A$ requires a recursive call. If $A$ is the entry point of the query, we ignore it — it gets special treatment later. If $needed(A)$ is empty, we can simply prune $A$. Otherwise, we will need to ask a subquery to obtain a distribution over $needed(A)$. For now, we assume that $Y.A$ has no asserted value in the knowledge base, so that the value of $Y.A$ is some unnamed instance $Z$ of class $T(A)$. Since the model of $Z$ is the same as that of every other unnamed instance of $T(A)$, we can ask the subquery on the class object $T(A)$. We therefore make a call to **SolveQuery**, in which the target is $T(A)$ and the set of outputs is $needed(A)$. In addition, if $A$ has an inverse $B$, the entry point is $B$, otherwise there is no entry point. The call to **SolveQuery** will return a set of inputs $\tau_A$, and a conditional probability distribution over $needed(A)$ given $\tau_A$. We treat the inputs $\tau_A$ to $A$ in the same way as the parents of a simple slot, using **GetChainNode**. We set the range of $\nu(A)$ to be $\times_{\sigma \in needed(A)} \text{Dom}(\sigma)$, and set the CPD of $\nu(A)$ to be that returned by the recursive call.

When we have finished processing all of the attributes, we can fill in the CPDs for the projection nodes. Each such node $\nu(\sigma)$ represents a component of the value of a complex attribute $A$. We could not specify the CPDs for these nodes at the time they were created, since we did not yet know the range of $\nu(A)$. Once all the nodes have been created, we simply set the CPD of $\nu(\sigma)$ to implement the projection function from $\times_{\sigma \in needed(A)} \text{Dom}(\sigma)$ onto $\sigma$.

At this point, we have almost built the complete network for solving the query. Recall that we have not yet processed the entry point $E$. The node $\nu(E)$ is the *input node*, representing the input of the query. We set the range of $\nu(E)$ to be $\prod_{\tau \in needed(E)} \text{Dom}(\tau)$. The node $\nu(E)$ has no parents and no CPD. We are now done building the local BN for the object $Y$. If the knowledge base asserts a value $v$ for a simple attribute $A$ of $Y$, we assert the value of $\nu(A)$ to be $v$ as evidence in the network. We then use a standard BN algorithm to compute the conditional probability of the output node given the input node, and return this conditional probability, along with the optional set of inputs $needed(E)$.

To summarize, let us consider how our algorithm exploits the two types of structure described in [KP97]. Each recursive call computes a distribution over the interface between two related objects. The algorithm exploits the fact that all the internal details of the callee are encapsulated from the caller by the interface. Much of the work of the algorithm, in particular maintaining the $needed()$ sets and returning the set of inputs $\tau$, is concerned with computing these interfaces on the fly.

As for exploiting the recurrence of the same type of object many times in the model, observe that different calls to **SolveQuery** with the same set of arguments will always return the same value. In order to reuse computation between different objects of the same type, we maintain a cache, in-dexed by the three arguments to **SolveQuery**. Note that we cannot reuse computation between different queries on the same object, because they may generate very different computations down the line. However, if the two queries are similar, many of the recursive computations they generate will be the same, and we will be able to reuse those.

### 4.2 Dealing with instances

If an instance has a value asserted for one of its attributes, we can no longer use the generic class model for that instance. In addition, if one instance is related to another, the internals of the two instances are not necessarily encapsulated from each other. Consider, for example, three instances $I$, $J$ and $K$, such that $I.A = J$, $I.B = K$, and $J.C = K$. In this case $K$ is not encapsulated from $I$ by $J$. Hence, the interface between $I$ and $J$ does not separate the internals of $I$ from the internals of $J$. When answering a query on $I$, we cannot simply perform a recursive call to obtain a distribution over the interface between $I$ and $J$, as we would lose the fact that the internals of $I$ and $J$ may be correlated by their mutual dependence on $K$. A possible solution to this problem is to include $K$ in the interface between $I$ and $J$. However, including entire objects in an interface creates huge interfaces, and defeats the purpose of using an object-based inference algorithm.

In order to deal with this issue, we create a new top-level object $\mathcal{T}$ in the knowledge base. This object contains an attribute $I{::}A$ for every named instance $I$ and each of its attributes $A$. If $A$ is simple, $\text{Dom}(I{::}A) = \text{Dom}(A)$; if it is complex, $T(I{::}A) = T(A)$. If the KB asserts a value for $I.A$, $I{::}A$ has the same value. Every user query will be directed through the top-level object. More precisely, a user query will have the form $\boldsymbol{I.\sigma} = I_1.\sigma_1, \ldots, I_n.\sigma_n$, where each $I_j$ is an instance (not necessarily distinct), and $\sigma_j$ is an attribute chain on $I_j$. The query is answered using a call **SolveTopLevel**($\mathcal{T}$, $\boldsymbol{I.\sigma}$). Since this is the top-level query, there is no entry point, and we will simply return a distribution over $\boldsymbol{I.\sigma}$.

**SolveTopLevel** is very similar to **SolveQuery**, so we omit the details. The main difference is in the way attribute chains are treated. On the top level, all attribute chains are attached to an instance. This is true both for the attribute chains required in the query output, and for the parents of any top-level attribute $I{::}A$. We replace **GetChainNode** with a function **GetTopLevelChainNode** that takes two arguments: an instance and an attribute chain. This function behaves similarly to **GetChainNode**, except for one situation. If the chain $\sigma$ is of the form $A.\rho$, and $I.A = J$, then $I.A.\sigma = J.\rho$. The algorithm eliminates this step of indirection, and continues to follow the rest of the chain.

### 4.3 Multi-valued attributes & structural uncertainty

We first show how to perform inference with multi-valued attributes with no number uncertainty. Let $A$ be a multi-valued complex attribute with $n$ values. One way to perform inference with multi-valued attributes is to replace $A$ with an array of $n$ single-valued attributes $A_1, \ldots, A_n$. Now consider a quantifier $Q = \#(A.\rho = v)$. The value of $Q$ depends on the values of $A_1.\rho, \ldots, A_n.\rho$. We therefore introduce a projection node for each of the $A_i.\rho$, and $\rho$ is added to $needed(A_i)$. The CPT for $\nu(Q)$ is defined to count the number of $\nu(A_i.\rho)$ whose value is $v$.

The problem with this approach is that $\nu(Q)$ has $n$ parents, so its CPT has size exponential in $n$. A better solution uses the fact that the value of $\nu(Q)$ does not depend directly on each of the individual $\nu(A_i.\rho)$, but only on the number of them that have the value $v$. Therefore, we do not need to introduce the $\nu(A_i.\rho)$ explicitly. Given the probability $p$ that a single $A_i.\rho$ has the value $v$, the probability that $k$ out of $n$ have the value $v$ is given simply by the binomial term $\binom{n}{k} p^k (1-p)^{n-k}$. We can compute the CPT for $\nu(Q)$ in time $O(n^2)$, using a recurrence relation. Let $P_m(Q = k)$ denote the probability that $k$ out of the first $A_1.\rho, \ldots, A_m.\rho$ have value $v$. Then $P_0(Q = 0) = 1$, and $P_{m+1}(Q = k) = (1-p)P_m(Q = k) + pP_m(Q = k-1)$.

If there are multiple quantifiers $Q_1, \ldots, Q_\ell$, where $Q_i = \#(A.\rho_i = v_i)$, the situation is a little more complicated, but similar. Since, for a particular value of $A$, the $\rho_j$ may not be independent of each other, the different quantifiers are also not independent of each other, and we need to compute a joint distribution over all of them. We begin by making a recursive call to **SolveQuery** for $A$, where $needed(A) = \{\rho_1, \ldots, \rho_\ell\}$. This gives us a distribution over the values of the $\rho_j$ for a single value of $A$. We can encode the contribution of a single value of $A$ to the quantifiers in a vector of length $\ell$, in which the $j$-th component is 1 if $A_i.\rho_j = v_j$, 0 otherwise. The distribution over the $2^\ell$ such contributions can be computed from the result of the recursive call to **SolveQuery**. Let $p_{\vec{c}}$ denote the computed probability of the contribution $\vec{c}$. Similarly, each joint value of the quantifier variables can be encoded by a vector of $\ell$ components, where each component is between 0 and $n$. We can then use a similar recurrence relation to the one above to compute the joint distribution over the values of all the quantifiers:

$$P_{m+1}(\boldsymbol{Q} = \vec{k}) = \sum_{\substack{\vec{c} \in \{0,1\}^\ell, \vec{k'} \in \{0,\ldots,m\}^\ell \\ \vec{k} = \vec{c} + \vec{k'}}} p_{\vec{c}} \cdot P_m(\boldsymbol{Q} = \vec{k'})$$

We need to compute $O((n+1)^{\ell+1})$ summations, and each one involves $O(2^\ell)$ terms, so the time to compute the joint distribution over the quantifiers is $O((2(n+1))^{\ell+1})$.

This computation is accomplished within the framework of **SolveQuery** by setting the range of $\nu(A)$ to be the set of possible values of the quantifiers. In the above discussion, we assumed that $A$ has no inputs. If it does, we of course have to perform the above computation for each value of the inputs. We then set the CPT for $\nu(A)$ to choose the computed joint probability distribution over the $Q_j$ given each value of the inputs of $A$. Finally, the CPT for $\nu(Q_j)$ simply computes a projection from the range of $\nu(A)$.

As mentioned earlier, we cannot deal with structural uncertainty by reasoning about all possible structures. Instead, we need to exploit the fact that many of the structural variables do not interact. Each of the complete structural hypotheses can be decomposed into many independent or conditionally independent sub-hypotheses. For example, the numbers of units of different types in a battery may all be independent of each other given the country to which the battery belongs. This type of reasoning about the conditional independence between different hypotheses is ideally performed in a BN. We need to express all possible structures within a single network, so that the BN inference algorithm can exploit these independencies.

This can be accomplished quite simply for number uncertainty in the above framework. We add $\nu(\#A)$ as a parent of $\nu(A)$. In the above recurrence relations, $P_m(Q = k)$ is the probability that the contributions of the first $m$ values of $A$ to the set of quantifiers will total $k$. This is exactly the probability we need to use if the value of $\nu(\#A) = m$. So the entire CPT for $\nu(A)$ with number uncertainty can be computed in the same time as above.

We deal with reference uncertainty as follows. Let $A$ be a complex attribute with reference uncertainty. For each value $v$ in the range of $\mathcal{R}(A)$, we create an attribute $A_v$, and a corresponding BN node $\nu(A_v)$. If $v$ is equal to the type $C$, we set the type of $A_v$ to be $C$. This operation ensures that $A_v$ will later be processed correctly, as a generic attribute of type $C$. If $v$ is equal to the instance $I$, we set the value of $A_v$ to be $I$. In this case, $A_v$ will be processed as a named instance. While we do not know the actual value of $A$, introducing all these nodes into the network accounts for all possible hypotheses over its value. We can now deal with dependencies on $A$. Consider a projection node $\nu(A.\rho)$. We introduce a set of projection nodes $\nu(A_v.\rho)$ for each value $v$ of $\mathcal{R}(A)$. We make $\nu(A.\rho)$ depend on all the $\nu(A_v.\rho)$ as well as on $\mathcal{R}(A)$, and set its CPD to select the value of the parent specified by $\mathcal{R}(A)$. We can think of the CPD of $\nu(A.\rho)$ as implementing a multiplexer, with selector $\mathcal{R}(A)$. (See [BFGK96] for a similar construction.)

## 5 Experimental Results

We have implemented the SPOOK system for representing models in the SPOOK language, and performing inference on these models. At the core of the system is a module containing the data structures necessary to represent the SPOOK data model. On top of this module is a user in-terface (see Figure 1(b)) in which the user can create class and instance objects, build probability models, observe values and query probabilities. SPOOK models can also be stored in an external knowledge server such as Ontolingua. SPOOK communicates with the knowledge server using the *OKBC protocol* [CFF+98], a generic protocol for communication between knowledge-based systems. Inference can be performed in SPOOK either by using the KBMC algorithm, or the object-based inference algorithm described in Section 4. Both inference methods use the same home-grown underlying BN inference engine.

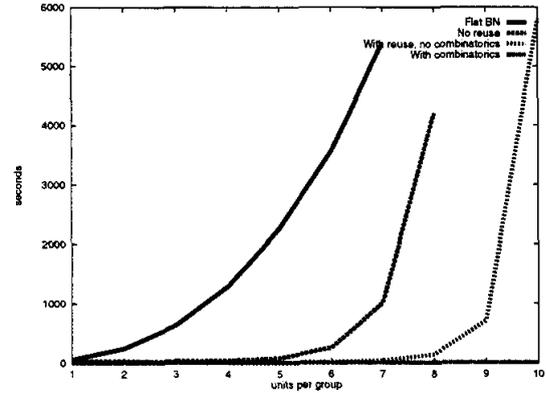

Figure 2: Experimental results

In our experiments, we compared the performance of the object-based algorithm with the KBMC algorithm on models of different sizes. Each model consists of a single battalion with four batteries, each containing 11 groups of different kinds with the number of units in each group varying from 1 to 9. The model also contains objects for the environment, location and weather, as described in Section 3. The size of the constructed BN grows linearly in the number of units per group, and varies from 750 to 5500 nodes.

In order to measure separately the benefits from exploiting interfaces and from reusing computation, we tried two different versions of the object-based algorithm, with and without reuse. We also compared the naive and combinatoric approaches to dealing with multivalued attributes described in Section 4.3. Figure 2 shows the running time in seconds of the different algorithms. From the graph, we see that all versions of the object-based algorithm outperform the KBMC algorithm by a large margin, and that the algorithm with reuse outperforms the algorithm without reuse. For example, with four units per group, the object-based algorithm with reuse takes 9 seconds, without reuse takes 46 seconds, while the KBMC algorithm takes 1292 seconds. In addition, we see that for the combinatoric approach to number uncertainty, the inference cost hardly increases with the number of units per group (its performance curve is just above the $x$-axis). Even with 10 units per group, the cost of inference topped out at 11 seconds, whereas for the naive approach we see an exponential blowup as we increase the number of units per group.

The reason for the great disparity between the inference times for the flat BN and for the object-based algorithm without reuse, is that the BN reasoning algorithm is failing to find optimal junction trees in the flat BN. The largest clique constructed for the flat BN contains 18 nodes, whereas the largest clique over all of the local BN computations for the structured algorithm contains only 8 nodes. The BN inference engine uses the standard minimum discrepancy triangulation heuristic to construct the junction tree. We see that at least for a standard BN implementation, exploiting object structure and the small interfaces between objects is vital to scaling up inference. While algorithms do exist for computing optimal triangulations [SG97], most implementations of Bayes nets do not use them, and these algorithms do not address the issue of reuse.

## 6 Discussion

An alternative approach to ours to modeling large, complex domains probabilistically is the *network fragments* approach of Laskey and Mahoney [LM97]. They provide network fragments for different aspects of a model, and operations for combining the fragments to produce more complex models. Network fragments exploit the same types of domain structure as do OOBNs. Because they allow complex fragments to be constructed out of simpler ones, they allow models to be composed hierarchically. Similarly, because they allow the same fragment to be reused multiple times, they exploit the redundancy in the domain.

The main difference between the two approaches is that ours focuses on building structured models, while theirs focuses on exploiting the domain structure for the knowledge engineering process, but the constructed models themselves are unstructured. An analogy from programming languages is that network fragments are like macros, which are preprocessed and substituted into the body of a program before compilation. SPOOK class models, on the other hand, are like defined functions, which become part of the structure of the compiled program. The advantages of the two approaches are comparable to those of their programming language analogues. Network fragments, like macros, have the advantage of flexibility, since no assumptions need be made about the relationship between combined fragments. For example, the restriction of OOBNs to part-of relationships was never an issue in the network fragment approach. The SPOOK language, on the other hand, provides a stricter, more semantic approach to combining models. Like structured programming languages, it allows strong type-checking in the definition of models.

The most important advantage of the SPOOK approach is that the models are themselves structured. The domain structure can then be exploited for efficient inference, as explained in Section 4. As our experimental results in Section 5 show, exploiting the domain structure can lead to great computational savings. In addition, because the domain structure is an explicit part of the model, we can now integrate uncertainty over the structure directly into the probability model.

SPOOK provides a bridge between probabilistic reasoning and traditional logic-based knowledge representation. Because it utilizes explicit notions of objects and the relationships between them, SPOOK is able to incorporate and augment the relational models used in many KR systems. This capability is enhanced by the ability of SPOOK to communicate with such systems through the OKBC protocol.

Our experiences with SPOOK are encouraging. Our hypothesis that exploiting the object structure of a domain can help both in knowledge representation and inference seems to be correct. Of course, we have only worked with one domain, and it remains to be seen if the advantages carry over to other domains. If they do, perhaps the door will be opened to a wide range of new and more complex applications of Bayesian network tecnhology.

### Acknowledgements

Grateful thanks to Suzanne Mahoney, KC Ng, Geoff Woodward and Tod Levitt of IET Inc. for their battlespace models; to Uri Lerner, Lise Getoor and all the Phrog hackers; to Barbara Engelhardt and Simon Tong for help with the knowledge engineering; and to Jim Rice for help with integrating with Ontolingua. This work was supported by ONR contract N66001-97-C-8554 under DARPA's HPKB program, by DARPA contract DACA76-93-C-0025 under subcontract to Information Extraction and Transport, Inc., and through the generosity of the Powell foundation.

## References

[BFGK96] C. Boutilier, N. Friedman, M. Goldszmidt, and D. Koller. Context-specific independence in Bayesian networks. In *Proc. UAI*, 1996.

[CFF+98] V.K. Chaudhri, A. Farquhar, R. Fikes, P. Karp, and J.P. Rice. A programmatic foundation for knowledge base interoperability. In *Proc. AAAI*, 1998.

[KP97] D. Koller and A. Pfeffer. Object-oriented Bayesian networks. In *Proc. UAI*, 1997.

[KP98] D. Koller and A. Pfeffer. Probabilistic frame-based systems. In *Proc. AAAI*, 1998.

[LM97] K. Laskey and S.M. Mahoney. Network fragments: Representing knowledge for constructing probabilistic models. In *Proc. UAI*, 1997.

[ML96] S.M. Mahoney and K. Laskey. Network engineering for complex belief networks. In *Proc. UAI*, 1996.

[SG97] K. Shoikhet and D. Geiger. A practical algorithm for finding optimal triangulations. In *Proc. AAAI*, 1997.

[WBG92] M.P. Wellman, J.S. Breese, and R.P. Goldman. From knowledge bases to decision models. *The Knowledge Engineering Review*, 7(1):35–53, 1992.